%% file: main.tex
\documentclass[10pt,twocolumn,letterpaper]{article}

\usepackage[pagenumbers]{cvpr} %

\input{preamble}

\definecolor{cvprblue}{rgb}{0.21,0.49,0.74}
\usepackage[pagebackref,breaklinks,colorlinks,citecolor=cvprblue]{hyperref}
\usepackage{multirow}
\usepackage{pifont}
\usepackage{ftnright}

\makeatletter
\g@addto@macro\normalsize{%
  \setlength\abovedisplayskip{4pt}
  \setlength\belowdisplayskip{3pt}
  \setlength\abovedisplayshortskip{4pt}
  \setlength\belowdisplayshortskip{3pt}
}
\makeatother

\makeatletter
\def\footnoterule{\kern-3\p@
  \hrule \@width 2in \kern 2.6\p@} %
\makeatother

\newcommand{\cmark}{\color{green}{\ding{51}}}
\newcommand{\xmark}{\color{red}{\ding{55}}}

\def\paperID{17} %
\def\confName{3DV\xspace}
\def\confYear{2024\xspace}

\title{Dynamic 3D Gaussians:\\Tracking by Persistent Dynamic View Synthesis\vspace{-1.3em}}

\author{
Jonathon Luiten$^{1,2}$%
\quad
Georgios Kopanas$^3$%
\quad
Bastian Leibe$^2$%
\quad
Deva Ramanan$^1$ \\[0pt]
{\small
$^1\hspace{1pt}$Carnegie Mellon University, USA%
\quad 
$^2\hspace{1pt}$RWTH Aachen University, Germany
\quad 
$^3\hspace{1pt}$Inria \& Université Côte d’Azur, France} \\[0pt]
{\tt\small luiten@vision.rwth-aachen.de}\\
}

\begin{document}

\twocolumn[{%
\vspace{-4.4em}
\maketitle
\vspace{-1.2em}

\begin{center}
    \centering 
    
    \vspace{-0.3in}
    \includegraphics[width=0.99\linewidth]{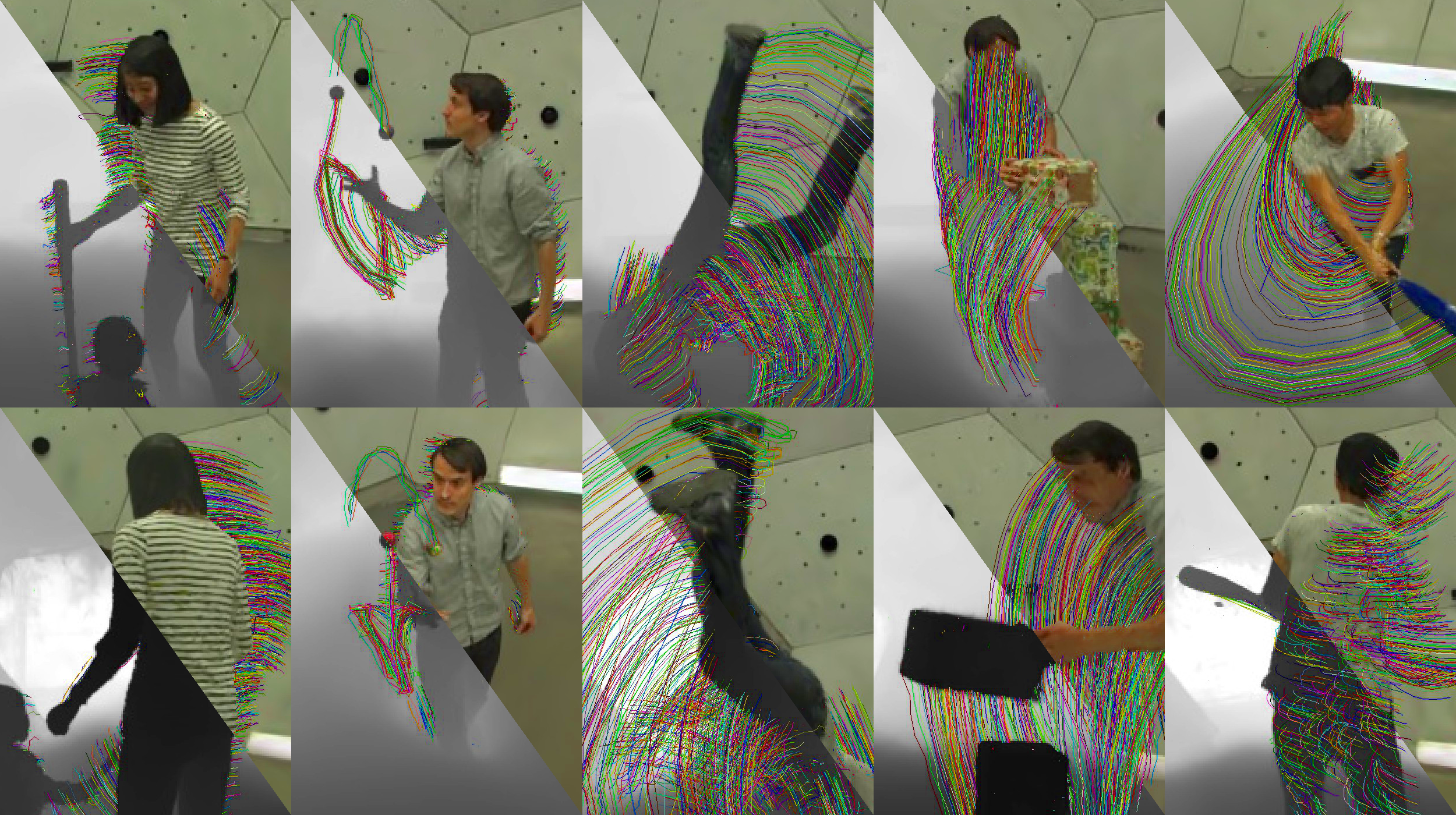}
    \captionof{figure}{\textbf{Persistent Dynamic Novel-View Synthesis and Tracking Results.} Novel-view (unseen) renders of color images and depth maps across 5 scenes (columns) and 2 views (rows) at the same timestep. Each scene is parameterized by 200-300k Dynamic 3D Gaussians which move over time. We render (with occlusions) the 3D trajectories of 2.5\% of these over the last 15 timesteps (0.5s). \href{https://dynamic3dgaussians.github.io/}{[Videos]} 
    }
    \label{fig:teaser}
\end{center}
}]

\input{sec/0_abstract}    
\input{sec/1_intro}
\input{sec/2_related}

\input{sec/3_method}
\input{sec/4_experiments}
\input{sec/5_conclusion}
{
    \small
    \bibliographystyle{ieeenat_fullname}
    \bibliography{main}
}

\end{document}

%% file: preamble.tex
\usepackage[dvipsnames]{xcolor}

\newcommand{\PAR}[1]{\noindent {\bf #1~}}

%% file: sec/0_abstract.tex
\begin{abstract}
\vspace{-1.5em}
We present a method that simultaneously addresses the tasks of dynamic scene novel-view synthesis and six degree-of-freedom (6-DOF) tracking of all dense scene elements. We follow an analysis-by-synthesis framework, inspired by recent work that models scenes as a collection of 3D Gaussians which are optimized to reconstruct input images via differentiable rendering. To model dynamic scenes, we allow Gaussians to move and rotate over time while enforcing that they have persistent color, opacity, and size. By regularizing Gaussians' motion and rotation with local-rigidity constraints, we show that our Dynamic 3D Gaussians correctly model the same area of physical space over time, including the rotation of that space. Dense 6-DOF tracking and dynamic reconstruction emerges naturally from persistent dynamic view synthesis, without requiring any correspondence or flow as input. We demonstrate a large number of downstream applications enabled by our representation, including first-person view synthesis, dynamic compositional scene synthesis, and 4D video editing. \footnote{\href{https://dynamic3dgaussians.github.io/}{Project Website.}}
\end{abstract}

%% file: sec/1_intro.tex
\vspace{-6pt}
\section{Introduction}
\vspace{-3pt}
\label{sec:intro}

Persistent dynamic 3D world modeling would be transformative for both discriminative and generative artificial intelligence.  On the discriminative side, this would enable a metric-space reconstruction of every part of the scene over time. Modeling where everything currently is, where it has been, and where it is moving, is crucial for many applications including robotics, augmented reality and self-driving. In generative AI, such models could enable new forms of content creation such as easily controllable and editable high resolution dynamic 3D assets for use in movies, video games or the meta-verse. Many such applications require scalable approaches that can be run on high-resolution imagery in real-time. Thus far, no approach has been able to produce photo-realistic reconstructions of arbitrary dynamic scenes with highly-accurate tracks and visually-appealing novel-views, all while being able to be trained quickly and rendered in real-time. 

In this paper we present such an approach by simultaneously tackling the discriminative tasks of dynamic 3D scene reconstruction and dense non-rigid long-term 6-DOF scene-tracking, while addressing the generative task of dynamic novel-view synthesis. We formulate both of these tasks in an analysis-by-synthesis framework, \ie, we build a persistent dynamic 3D representation of the moving scene that is consistent with all the input observations (images from different timesteps and cameras) and from which tracking emerges as a product of correctly modelling the underlying scene with physically plausible spatial consistency priors.

3D Gaussian Splatting~\cite{kerbl20233d} has recently emerged as a promising approach to modelling 3D static scenes. It represents complex scenes as a combination of a large number of coloured 3D Gaussians which are rendered into camera views via splatting-based rasterization. The positions, sizes, rotations, colours and opacities of these Gaussians can then be adjusted via differentiable rendering and gradient-based optimization such that they represent the 3D scene given by a set of input images. In this paper we extend this approach from modelling only static scenes to dynamic scenes. 

Our key insight is that we restrict all attributes of the Gaussians (such as their number, color, opacity, and size) to be the same over time, but let their position and orientation vary.
This allows our Gaussians to be thought of as a {\em particle-based} physical model of the world, where oriented particles 
undergo rigid-body transformations over time. In order to reconstruct such particles from raw camera imagery, we exploit the property that at any time, they can be efficiently splatted onto to any camera viewpoint, allowing them to be optimized with an image reconstruction loss. %
Crucially, particles allow us to operationalize {\em physical priors} over their movement that act as regularizers for the optimization:
a local rigidity prior, a local rotational-similarity prior, and a long-term local isometry prior. These priors ensure that local neighborhoods of particles move approximately rigidly between timesteps, and that nearby particles remain closeby over all timesteps. %

Previous approaches to neural reconstruction of dynamic scenes can be seen as either Eulerian representations that keep track of scene motion at fixed grid locations~\cite{fridovich2023k, cao2023hexplane, turki2023suds} or Lagrangian representations where an observer follows a particular particle through space and time. We fall in the latter category, but in contrast to prior point-based representations~\cite{abou2022particlenerf, zhang2022differentiable}, we make use of oriented particles that allow for richer physical priors (as above) and that directly reconstruct the 6-DOF motion of all 3D points, enabling a variety of
downstream applications (see Fig.~\ref{fig:rot} and~\ref{fig:creative}).

We perform experiments using synchronized multi-view video (27 training cameras, 4 testing cameras) from the CMU Panoptic Studio dataset \cite{joo2015panoptic}. Our approach achieves 28.7 PSNR on dynamic novel view rendering while rendering at 850 FPS. It is trained on 150 timesteps with 27 training cameras in each timestep in only 2 hours on a single RTX 3090 GPU. Furthermore, our approach results in accurate metric 3D dense non-rigid long-term scene tracking with an average $L2$ error of only 2.21cm in 3D over 150 timesteps, and having an average of 1.57 normalized-pixel error on 2D tracking metrics, which is an order of magnitude (10x) better than previous state-of-the-art. Our method is also able to track the rotation of every 3D point in space, enabling full 6-DOF dense scene tracking. We show visual results in Fig.~\ref{fig:teaser}. %

An remarkable feature of our approach is that tracking arises exclusively from the process of rendering per-frame images. No optical flow, pose skeletons, or any other form of correspondence information is given as input. Due to its persistent and naturally decomposable nature, Dynamic 3D Gaussians are naturally amenable to a number of creative scene editing techniques such as propagating edits over all timesteps, adding or removing dynamic objects to a scene, or having cameras follow scene elements, as seen in Fig~\ref{fig:creative}. Furthermore, the extremely fast rendering and training time make them much easier to work with than previous approaches for dynamic reconstruction~\cite{joo2014map}, and enable real-time rendering applications.

%% file: sec/2_related.tex
\vspace{-6pt}
\section{Related Work}
\vspace{-3pt}
\label{sec:related}

In this work, we are interested in solving the combination of dynamic novel-view synthesis, long-term point-tracking and dynamic reconstruction in a unified analysis-by-synthesis framework. 

\textbf{Dynamic Novel-View Synthesis.} The general field of novel-view synthesis exploded in popularity with the release of NeRF~\cite{mildenhall2020nerf} in 2020. Since then there have been a large number of `dynamic NeRF' papers extending the idea of fitting a 3D radiance field to the 4D dynamic domain. Most of these use monocular video as input, whereas we focus on using multi-camera capture. Typically these methods belong to one of the following 5 categories: 

(a) Methods that fit a separate representation per timestep, and cannot model correspondence over time~\cite{xian2020space,bansal2023neural}. 

(b) Methods that represent the scene using an Eulerian representation on a 4D space-time grid, often with various grid decompositions for efficiency such as planar decomposition or hash functions~\cite{fridovich2023k, turki2023suds, cao2023hexplane}. Such Eulerian approaches also do not give rise to correspondence over time.

(c) Methods that represent the 3D scene in a canonical timestep and use a deformation field to warp this to the rest of the timesteps~\cite{park2021nerfies, pumarola2021d, park2021hypernerf, li2021neural, du2021neural, yang2022banmo, liu2023robust, wang2023tracking}. Such methods naturally result in one-directional backward correspondences between each frame and the reference frame, but by default, not the forward correspondences which are required to generate correspondences between any two timesteps. \cite{gao2022monocular} advocates for expensive ad hoc root-finding to determine and evaluate such correspondences. A number of methods specifically parameterize their warp-fields in a way that is easily invertable such as linear-blend-skinning~\cite{yang2022banmo, song2023total}, or reversible neural networks~\cite{wang2023tracking}, in order to obtain correspondences. While this often works quite well~\cite{yang2022banmo, song2023total, wang2023tracking}, requiring a single canonical view for a scene greatly restricts the dynamic representation ability. 

(d) Template guided methods~\cite{li2022tava, weng2022humannerf, isik2023humanrf}, which model dynamic scenes in restricted environments where the motion can be modelled by a predefined template e.g.~a set of human-pose skeleton transformations. This approach often requires a-priori knowledge of what is to be reconstructed and thus is not a solution for general scenes.

(e) Point-based methods \cite{abou2022particlenerf, zhang2022differentiable}, which compared to all of the above categories, hold the most promise for representing dynamic scenes in a way where accurate correspondence over time can emerge due to their natural Lagrangian representation. However, these haven't achieved as much attention because the point-based rendering approaches haven't worked as well as MLP \cite{mildenhall2020nerf} or grid based approaches \cite{mueller2022instant}.

While a number of view-synthesis works have used Gaussians for static scenes \cite{kerbl20233d, keselman2022fuzzy, wang2022voge}, to the best of our knowledge we are the first to use them to reconstruct dynamic scenes. We build our dynamic Gaussian renderer upon the the static renderer `3D Gaussian splatting' \cite{kerbl20233d} which is currently the state-of-the-art static scene reconstruction algorithm in terms of both accuracy and speed.

Other than rendering accuracy and speed, modeling the dynamic world with Gaussians has a distinct advantage over points as Gaussian's have a notion of `rotation' so we can use them to model the full 6 degree-of-freedom (DOF) motion of a scene at every point and can use this to construct physically-plausible local rigidity losses.

\PAR{Long-Term Point Tracking.} Traditionally video tracking algorithms have been focused on tracking whole objects (as bounding boxes or segmentation masks) \cite{voigtlaender2019mots, Caelles_arXiv_2019, Kristan_2021_ICCV}, or tracking dense scene points but only between two timesteps (e.g. optical-flow / scene flow) \cite{beauchemin1995computation,vedula1999three}. Recently, a number of approaches have started tackling the task of long-term dense point tracking \cite{harley2022particle, zheng2023pointodyssey, doersch2022tap, doersch2023tapir, karaev2023cotracker, wang2023tracking}, where every pixel in a video needs to be tracked across every video frame. In this paper, we extend this long-range tracking task to 3D and evaluate it in a multi-camera capture setup. Most of these prior approaches \cite{harley2022particle, zheng2023pointodyssey, doersch2022tap, doersch2023tapir, karaev2023cotracker} are deep learning based and work by being trained on a large-dataset of ground-truth point tracks (often using synthetic training data). The most similar method to ours is OmniMotion \cite{wang2023tracking} which also fits a dynamic radiance field representation using test-time optimization for the purpose of long-term tracking. They focus on monocular video while we focus on multi-camera capture and as such we can reconstruct tracks in metric-3D while they produce a `pseudo-3D' representation. 
However, the largest and most significant difference between our method and OmniMotion is that they require dense optical-flow input between every pair of timesteps as an optimization target, which is incredibly expensive as it scales with the number of frames squared. The optical flow estimates already provide a (noisy and inconsistent) tracking result which is made consistent through dynamic modelling of the scene. In contrast, our method takes no correspondences at all as input and tracking emerges from fitting a persistent representation to the input frames along with physically-based priors.

\textbf{Dynamic Reconstruction}
In addition to the radiance-field based approaches listed above, a number of other approaches have tackled the task of `dynamic reconstruction' \cite{newcombe2015dynamicfusion, slavcheva2017killingfusion, joo2014map, prokudin2023dynamic, barsan2018robust, luiten2019track}. Such approaches either rely on the presence of accurate depth cameras \cite{newcombe2015dynamicfusion, slavcheva2017killingfusion}, assume ground-truth object point clouds as input \cite{prokudin2023dynamic}, or are heavily specific to certain domains e.g. reconstructing moving cars in driving scenes \cite{barsan2018robust, luiten2019track}. The closest approach to ours is \cite{joo2014map} which also predicts long-term 3D point tracks and also uses the Panoptic Studio data for evaluation. However this approach requires 480 input cameras (in contrast we use 27) and also requires as input pre-computed optical-flow correspondences, which it then lifts into 3D trajectories.

%% file: sec/3_method.tex
\vspace{-6pt}
\section{Method}
\vspace{-3pt}
\label{sec:method}

\PAR{Overview.}
Given a set of images from different timesteps and different cameras ($\mathcal{I}_{t, c}$), along with each camera's respective intrinsic ($K_c$) and extrinsic ($E_{t,c}$) matrices, our approach reconstructs the dynamic 3D scene ($\mathcal{S}$) observed by these cameras in a temporally persistent manner. 

This reconstruction is performed via test-time optimization and no further training data other than the test scene is used. The reconstruction is performed temporally online, \ie, one timestep of the scene is reconstructed at a time with each one being initialized using the previous timestep's representation. The first timestep acts as an initialization for our scene where we optimize all properties, and then fix all for the subsequent timesteps except those defining the motion of the scene. Each timestep is trained via gradient based optimization using a differentiable renderer ($\mathcal{R}$) to render the scene at each timestep into each of the training cameras. 
$$\mathcal{\hat{I}}_{t, c} = \mathcal{R}(\mathcal{S}_t, K_c, E_{t,c})$$
The renderings $\mathcal{\hat{I}}_{t, c}$ are compared to the input images $\mathcal{I}_{t, c}$, and the parameters of $\mathcal{S}$ are iteratively updated using automatic differentiation in order to decrease the error between $\mathcal{\hat{I}}_{t, c}$ and $\mathcal{I}_{t, c}$. After convergence, the representation $\mathcal{S}_t$ is a 3D reconstruction of the scene given by each of the training cameras $\{\mathcal{I}_{t, c}, K_c, E_{t,c}\}$ for this timestep. By choosing a suitable representation for $\mathcal{S}$ and applying physically-based regularization losses during optimization, we can ensure that all the $\mathcal{S}_t$ are temporally consistent with one another and that there exists a one-to-one correspondence between every 3D point in every timestep, along with their corresponding changes in 3D rotation. In this way, tracking emerges from persistent dynamic view synthesis.

\PAR{Dynamic 3D Gaussians.}
Our dynamic scene representation ($\mathcal{S}$) is parameterized by a set of Dynamic 3D Gaussians, each of which has the following parameters:
\begin{enumerate}[label=\arabic*),itemsep=2pt]
\item a 3D center for each timestep ($x_t$, $y_t$, $z_t$).
\item a 3D rotation for each timestep parameterized by a quaternion ($qw_t$, $qx_t$, $qy_t$, $qz_t$).
\item a 3D size in standard deviations (consistent over all timesteps) ($sx$, $sy$, $sz$)
\item a color (consistent over all timesteps) ($r$, $g$, $b$)
\item an opacity logit (consistent over all timesteps) ($o$)
\item a background logit (consistent over timesteps) ($bg$)
\end{enumerate}
\vspace{2pt}
This gives a total of $7t + 8$ parameters for each Gaussian. In our experiments, scenes are represented by between 200-300k Gaussians, of which only 30-100k usually are not part of the static background. While the code contains the ability to represent view-dependent color using spherical harmonics, we turn this off in our experiments for simplicity.

Each Gaussian can be thought of as softly representing an area of 3D physical space which is occupied by solid matter.
Each Gaussian influences a point in physical 3D space ($p$) according to the standard (unnormalized) Gaussian equation weighted by its opacity:
$$f_{i, t}(p) = \textrm{sigm}({o_i}) \exp\left( -\frac{1}{2} (p - \mu_{i,t})^T \Sigma_{i, t}^{-1} (p - \mu_{i,t}) \right)$$
Where ${\mu}_{i,t} = \begin{bmatrix} x_{i,t} &  y_{i,t} &  z_{i,t} \end{bmatrix}^T$ is the center of each Gaussian $i$ at timestep $t$, and $\Sigma_{i,t} = R_{i,t} S_i S_i^T R_{i,t}^T$ is the covariance matrix of Gaussian $i$ at timestep $t$, given by combining the scaling component $S_i = \text{diag}\left(\begin{bmatrix} sx_i & sy_i & sz_i \end{bmatrix}\right)$, and the rotation component $R_{i,t} = \textrm{q2R}\left(\begin{bmatrix} qw_{i,t} & qx_{i,t} & qy_{i,t} & qz_{i,t} \end{bmatrix}\right)$, where $\textrm{q2R}()$ is the formula for constructing a rotation matrix from a quaternion. $\textrm{sigm}()$ is the standard sigmoid function.

\begin{figure}
    \centering
    \includegraphics[width=1.0\columnwidth]{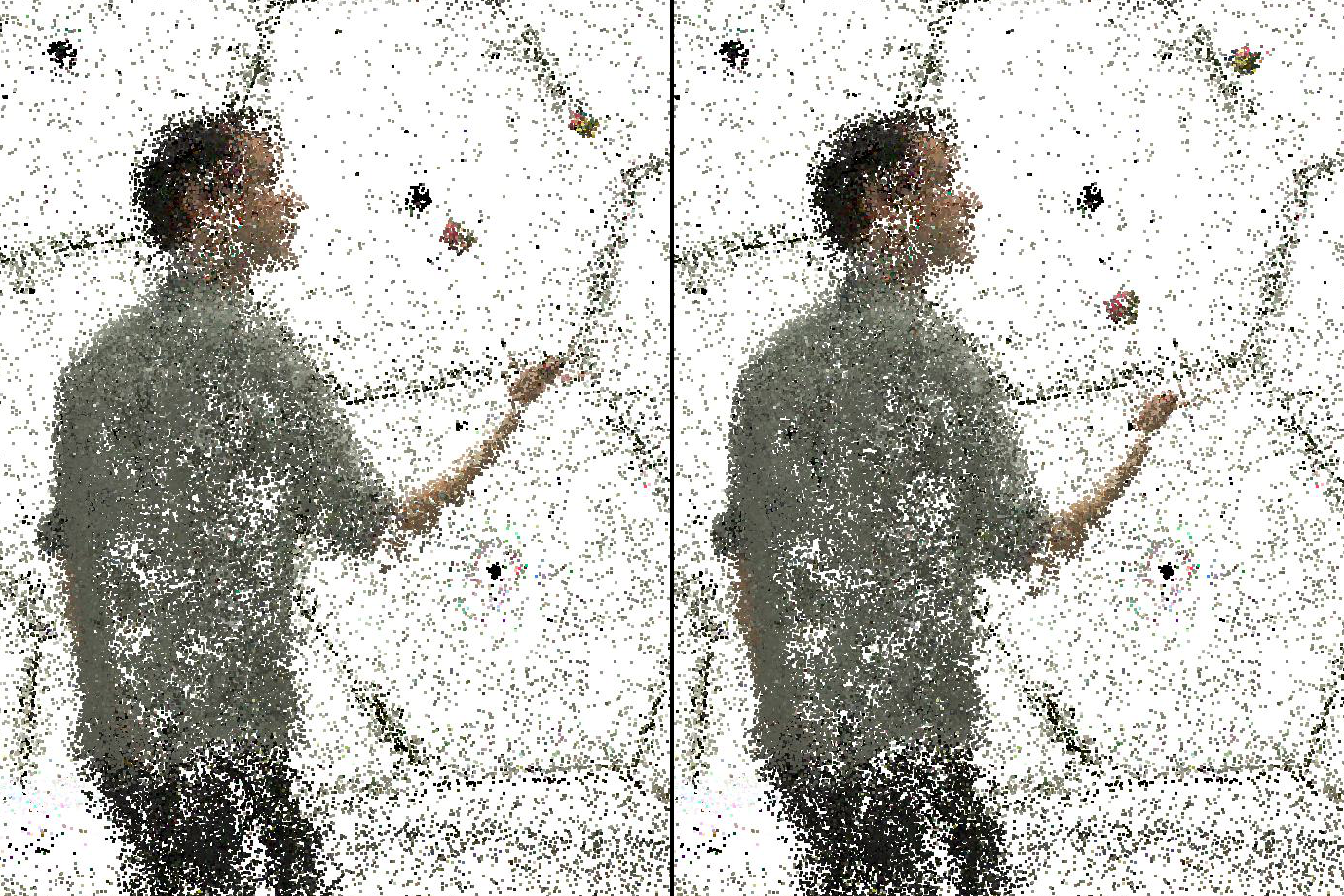}
    \caption{\textbf{Gaussian Centers.} Point-cloud of colored centers in contiguous timesteps, showing how they model scene geometry and move over time. \href{https://dynamic3dgaussians.github.io/}{[Videos]}}
    \label{fig:point_cloud}
\end{figure}

\begin{figure}
    \centering
    \includegraphics[width=1.0\columnwidth]{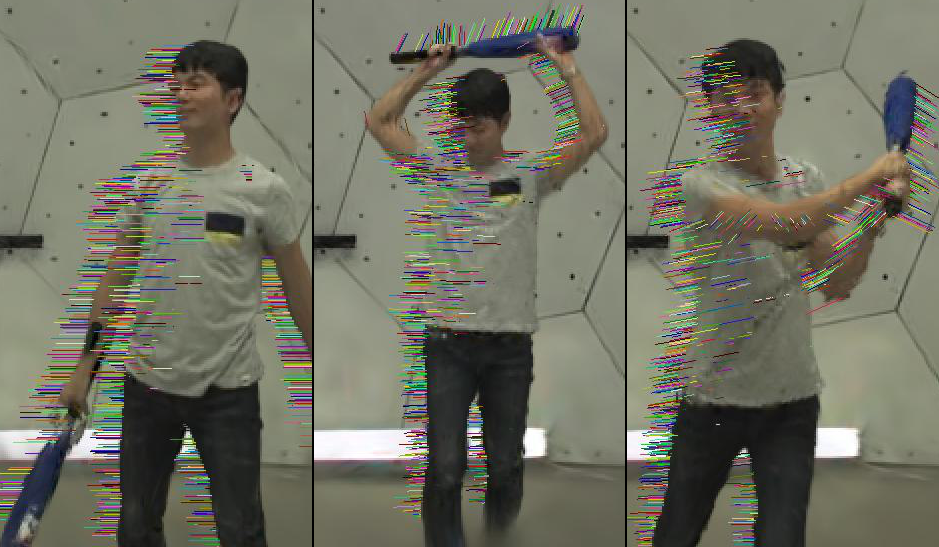}
    \caption{\textbf{Relative Rotation Tracking.} 1st panel: Left-facing coloured vectors are attached to 3\% of Gaussians in the first frame. 2nd and 3rd panels: These vectors move and rotate along with the Gaussians they are attached to, showing that our approach correctly models 6-DOF motion. \href{https://dynamic3dgaussians.github.io/}{[Videos]}}
    \label{fig:rot}
\end{figure}

Each Gaussian's influence ($f$) is both inherently local (being able to represent a small area of space), while also theoretically having infinite extent, such that gradients can flow to them even from a long distance, which is crucial for gradient-based tracking-by-differentiable rendering, as Gaussians which may currently be in the wrong 3D location need to get gradients pushing them towards moving to the correct 3D location through the differentiable renderer. The softness of this Gaussian representation also means that Gaussians typically need to significantly overlap in order to represent a physically solid object. As well as physical density, each Gaussian contributes its own color ($r$, $g$, $b$) to each of the 3D points it influences.

By fixing the size/opacity/color of the Gaussians across time, each Gaussian should represent the same physical aspect of space, even as this space dynamically moves through time. To represent this motion, each Gaussian has a center location and rotation that can move with time, enabling full dense non-rigid 6-DOF tracking of a whole scene. 

We visualize the trajectories of these Gaussians in Fig~\ref{fig:teaser}, the locations of the Gaussian's centers in Fig~\ref{fig:point_cloud}, and the change in rotation over time in Fig~\ref{fig:rot}. 

\PAR{Differentiable Rendering via Gaussian 3D Splatting.}
In order to optimize the parameters of our Gaussians to represent the scene, we need to render the Gaussians into images in a differentiable manner. In this work we use the differentiable 3D Gaussian renderer from \cite{kerbl20233d} and extend its use to dynamic scenes. This works by splatting 3D Gaussians into the image plane by approximating the projection of the integral of the influence function $f$ along the depth dimension of the 3D Gaussian into a 2D Gaussian influence function in pixel coordinates. The center of the Gaussian is splatted using the standard point rendering formula:
$$\mu^{\textrm{2D}} = K \left((E \mu) / (E \mu)_z\right)$$
where the 3D Gaussian center $\mu$ is projected into a 2D image by multiplication with the world-to-camera extrinsic matrix $E$, z-normalization, and multiplication by the intrinsic projection matrix $K$. The 3D covariance matrix is splatted into 2D using the formula from \cite{zwicker2001surface}:
$$\Sigma^{\textrm{2D}} = J E \Sigma E^T J^T$$
where $J$ is the Jacobian of the point projection formula above, i.e. $\partial \mu^{\textrm{2D}} / \partial \mu$.

The influence function $f$ can now be evaluated in 2D for each pixel for each Gaussian. The influence of all Gaussians on this pixel can be combined by sorting the Gaussians in depth order and performing front-to-back volume rendering using the Max \cite{max1995optical} volume rendering formula (the same as is used in NeRF \cite{mildenhall2020nerf}):
$$C_{\text{pix}} = \sum_{i \in \mathcal{S}} c_i f^{\textrm{2D}}_{i, \textrm{pix}} \prod_{j=1}^{i-1} (1 - f^{\textrm{2D}}_{j, \textrm{pix}})$$
where the final rendered color ($C_{\text{pix}}$) for each pixel is a weighted sum over the colors of each Gaussian ($c_i = \begin{bmatrix} r_i &  g_i &  b_i \end{bmatrix}^T$), weighted by the Gaussian's influence on that pixel $f^{\textrm{2D}}_{i, \textrm{pix}}$ (the equivalent of the formula for $f_i$ in 3D except with the 3D means and covariance matrices replaced with the 2D splatted versions), and down-weighted by an occlusion (transmittance) term taking into account the effect of all Gaussians in front of the current Gaussian. 

The implementation of \cite{kerbl20233d} uses a number of graphics and CUDA optimization techniques to achieve incredibly fast rendering speeds (\eg, $850$ FPS for our scenes), which therefore also enables very fast training.

\begin{figure}
    \centering
    \includegraphics[width=1.0\columnwidth]{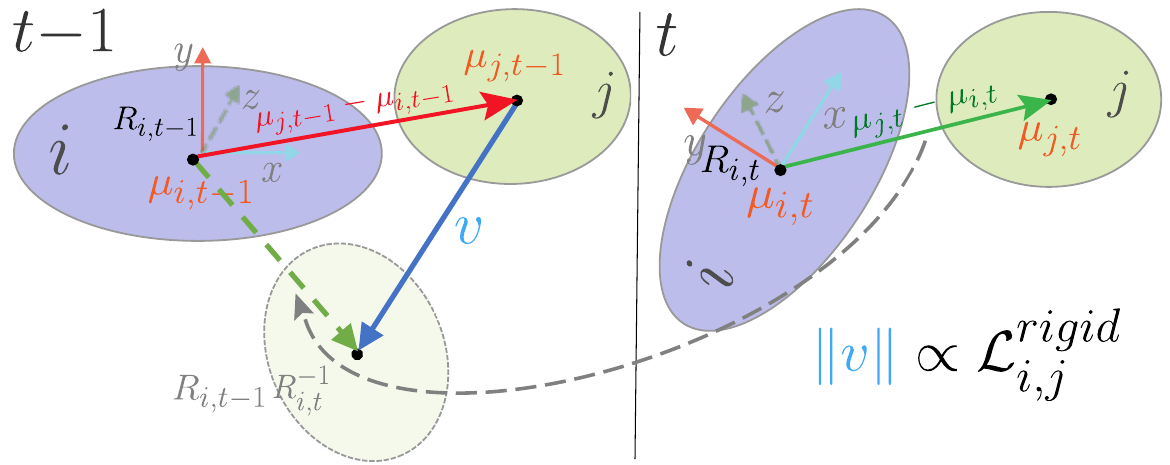}
    \caption{\textbf{Local Rigidity Loss.} For each Gaussian $i$, nearby Gaussians $j$ should move in a way that follows the rigid-body transform of the coordinate system of $i$ between timesteps.}
    \label{fig:loss}
\end{figure}

\PAR{Physically-Based Priors.} We find that just fixing the color, opacity and size of Gaussians is not enough on its own to generate long-term persistent tracks, especially across areas of the scene where there is a large area of near uniform colour. In such situation the Gaussians move freely around the area of similar colour as there is no restriction on them doing so. Since we are trying to model physically moving scenes, it makes sense to look to non-rigid physical modelling for inspiration on how to regularize the optimization procedure to be physically plausible and give correct long-term tracking results. We introduce three regularization losses, short-term local-rigidity $\mathcal{L}^{\text{rigid}}$ and local-rotation similarity $\mathcal{L}^{\text{rot}}$ losses and a long-term local-isometry loss.

The most important of these is the local-rigidity loss $\mathcal{L}^{\text{rigid}}$, defined as:
$$\mathcal{L}^{\text{rigid}}_{i, j} = w_{i,j} \left\| (\mu_{j,t-1} - \mu_{i,t-1}) - R^{}_{i,t-1} R^{-1}_{i,t} (\mu_{j,t} - \mu_{i,t}) \right\|_2$$
$$\mathcal{L}^{\text{rigid}} = \frac{1}{k |\mathcal{S}|} \sum_{i \in \mathcal{S}} \sum_{ j \in \text{knn}_{i;k}} \mathcal{L}^{\text{rigid}}_{i, j}$$
This states that, for each Gaussian $i$, nearby Gaussians $j$ should move in a way that follows the rigid-body transform of the coordinate system of $i$ between timesteps. See Fig~\ref{fig:loss} for a visual explanation. 

Since we are performing online optimization, all of $\mu_{i,t-1}, R^{}_{i,t-1}, \mu_{j,t-1}$ are fixed, and we are optimizing $\mu_{i,t}, R^{}_{i,t}, \mu_{j,t}$ to ensure that they match the values in $t-1$ up to the rigid body transformation defined by the change in Gaussian $i$'s own coordinate system. 
e.g. if $i$ rotates, then $j$ needs to translate (in it's own coordinate system) in a way that in equivalent to rotating around the center of $i$. This loss also applies the other way, forcing the rotation to match the translation, such that we obtain accurate rotation (6-DOF) tracking for every dense point in space, even though we only optimize to match the rendered images, which isn't possible with a point-based representation.

We restrict the set of Gaussians $j$ to be the k-nearest-neighbours of $i$ (k=20), and weight the loss by the a weighting factor for the Gaussian pair:
$$w_{i,j} = \exp\left( -\lambda_w \left\|\mu_{j,0} - \mu_{i,0} \right\|^2_2 \right)$$
which is an (unnormalized) isotropic Gaussian weighting factor. We set $\lambda_w$ to 2000, which gives a standard deviation of $\sim$2.2cm, and calculate this with the distance between the Gaussian centers in the first timestep and fix it over the rest of the timesteps. This results in the rigidity loss only being enforced locally, while still allowing global non-rigid reconstruction.

The rigidity loss is both necessary and adequate on it's own to achieve good results. Since the rigidity loss is applied on all points it is applied in both directions between any pair $i$ and $j$ and thus implicitly enforces $i$ and $j$ to have the same rotation, however we found better convergence if we explicitly force neighbouring Gaussians to have the same rotation over time:
$$\mathcal{L}^{\text{rot}} = \frac{1}{k |\mathcal{S}|} \sum_{i \in \mathcal{S}} \sum_{ j \in \text{knn}_{i;k}} w_{i,j} \left\| \hat{q}^{}_{j,t} \hat{q}^{-1}_{j,t-1} - \hat{q}^{}_{i,t} \hat{q}^{-1}_{i,t-1}  \right\|_2$$
where $\hat{q}$ is the normalized quaternion representation of each Gaussian's rotation, which enables smooth optimization. We use the same set of k-nearest-neighbours and weighting function as before.

We apply $\mathcal{L}^{\text{rigid}}$ and $\mathcal{L}^{\text{rot}}$ only between the current timestep and the directly preceding timestep, thus only enforcing these losses over short-time horizons. 
Which sometimes causes elements of the scene to drift apart, thus we apply a third loss, the isometry loss, over the long-term:
$$\mathcal{L}^{\text{iso}}\hspace{-0.5ex}=\hspace{-0.5ex}\frac{1}{k |\mathcal{S}|} \sum_{ i \in \mathcal{S}} \sum_{ j \in \text{knn}_{i;k}}\hspace{-1.5ex}w_{i,j} \left| \left\| \mu_{j,0} - \mu_{i,0} \right\|_2\hspace{-0.5ex}-\left\| \mu_{j,t} - \mu_{i,t} \right\|_2 \right|$$
This is a weaker constraint than $\mathcal{L}^{\text{rigid}}$ in that instead of enforcing the positions between two Gaussians to be the same it only enforces the distances between them to be the same.

\PAR{Optimization Details.} 
Each timestep of a scene is optimized one-at-a time. During the first timestep, all parameters of the Gaussians are optimized. After the first timestep the size, color, opacity, and background logit are fixed and only the position and rotation are updated. Thus, our approach can be seen as first performing static reconstruction of the first frame, followed by long-term dense 6-DOF tracking throughout the remaining frames.

Following \cite{kerbl20233d}, in the first timestep we initialize the scene using a coarse point cloud that could be obtained from running colmap, but instead we use available sparse samples from depth cameras. Note that these depth values are only used for initializing a sparse point cloud in the first timestep and are not used at all during optimization.

We use the densification from \cite{kerbl20233d} in the first timestep in order to increase the density of Gaussians and achieve a high quality reconstruction. For the rest of the frames the number of Gaussians is fixed and the densification is turned off.

We fit the 3D scene in the first frame for 10000 iterations (around 4 minutes) using 27 training cameras, where each iteration renders a single but complete image. For each timestep after that we use 2000 iterations (around 50 seconds), for a total of ~2 hours for 150 timesteps. At the beginning of each new timestep we initialize the estimated Gaussian center positions and rotation quaternion parameters by using forward estimate based on a velocity estimated from the current position minus the previous position, and do the same for the quaternion using normalized quaternions (e.g. we also re-normalize the quaternion representation). We find this to be quite important to getting good results. We also reset the Adam first and second order momentum parameters at the start of each timestep.

In our test scenes the subject's shirt colours (grey) are very similar to the background. We noticed that often the shirt was being mis-tracked as it was confused with the background, while more contrastive elements like pants and hair were being tracked correctly. To increase the contrast between foreground and background parts of the scene we also render a foreground/background mask and apply a background segmentation loss $\mathcal{L}^{Bg}$ against a pseudo-ground-truth background mask, which we can easily obtain by differencing with an image from the dataset where no foreground objects are presents. We also directly apply a loss that background points shouldn't move or rotate, and restrict the above rigidity, rotation and isometry losses to only operate over foreground points to improve efficiency. This also ensures that these losses are never enforced between foreground parts of the scene and the static floor.

Finally, since we are optimizing over 27 different training cameras, each of which has different camera properties such as white balance, exposure, sensor sensitivity, and color calibration, these factors contribute to variations in color representation across images. Thus we naively model these differences by simply optimizing a scale and offset parameter for each colour channel for each camera separately. We optimize these only over the first timestep and then fix these for the rest of the timesteps.

\PAR{Tracking with Dynamic 3D Gaussians.}
After we have trained our Dynamic 3D Gaussian scene representation we can use it to obtain dense correspondence for any point in a 3D scene. To determine the correspondence of any point in 3D space $p$ across timesteps, we can linearize the motion-space by simply taking the point's location in the coordinate system of the Gaussian that has the most influence $f(p)$ over this point (or the static background coordinate system if $f(p) < 0.5$ for all Gaussians). 
There is now a well-defined and invertible one-to-one mapping for all points in space across all timesteps giving dense correspondences.

We can use this same idea to track any 2D pixel-location from any input or novel view into any other timestep or view. To do so we first have to determine the 3D point corresponding to an input pixel. We can render out depth-maps for any view by using the Max \cite{max1995optical} rendering equation for Gaussians \cite{kerbl20233d} but replacing the colour component of each Gaussian with the depth of that Gaussian's center, as seen in Figure~\ref{fig:teaser}. A pixel's 3D location can be found via unprojection, the highest influence Gaussian determined, tracked and then projected into a new camera frame. 

%% file: sec/4_experiments.tex
\vspace{-6pt}
\section{Experiments}
\vspace{-3pt}
\label{sec:experiments}
\input{tables/cmu_panoptic}

\input{tables/particle-nerf}

\input{tables/ablations}

\PAR{Dataset Preparation.}
We prepare a dataset which we call \texttt{PanopticSports}. We take six sub-sequences from the sports sequence of the Panoptic Studio dataset \cite{joo2015panoptic}. Each of our six sequences contain interesting motions and objects which we name them after: \texttt{juggle}, \texttt{box}, \texttt{softball}, \texttt{tennis}, \texttt{football} and \texttt{basketball}. We produce visual results for 3 more sequences, \texttt{handstand}, \texttt{sway} and \texttt{lift} but don't include them in evaluation results as they don't have 3D track ground-truth available. For each sequence we obtain 150 frames at 30 FPS, from the set of the HD cameras. There are 31 cameras, which we split into 27 training and 4 testing cameras (cam 0, 10, 15 and 30 are test). Cameras are temporally aligned and have accurate intrinsics and extrinsics provided. The cameras are positioned roughly in a hemisphere around an area of interest in the middle of a capture studio dome. Fig~\ref{fig:teaser} shows an example. 

We undistort the images from each camera using the provided distortion parameters, and resize each image to be 640x360. We create an initial point cloud for the first timestep to initialize the Gaussians by taking the points from 10 available depth cameras, synchronized to the current timestep, subsample these depth maps by a factor of 2 in both dimensions and obtain an initial colour by projecting these points into the nearest training camera. Points that do not project into any training camera are discarded. We obtain pseudo-ground-truth foreground-background segmentation masks by simply doing frame-differencing between each frame and a reference frame from the dataset for each camera when no foreground objects are present. 

We prepare ground-truth trajectories for 2D and 3D tracking by taking the high-quality facial and hand key-point annotations that are available for the scene \cite{Simon_2017_CVPR}. For each person in each scene (four scenes have one person, two have two), we take one random face key-point and one random hand key-point from each hand. We manually verify the accuracy of these trajectories by projecting them into the images and viewing them, and remove three that are not accurate. This leaves us with a total of 21 ground truth 3D trajectories.

For 2D tracking, we use the camera-visibility labels to determine if the first point in each 3D trajectory is visible in each camera, and add these videos and projected points to our 2D tracking evaluation. Each 3D point is visible in around 18 cameras for a total of 371 2D ground-truth tracks.

\PAR{Evaluation Metrics.}
We evaluate novel-view synthesis on the hold-out 4 camera views across all 150 timesteps for the 6 sequences. We use the standard PSNR, SSIM and LPIPS metrics \cite{zhang2018perceptual, wang2004image}.
For 2D long-term point tracking we use the metrics from the recent point-odyssey benchmark \cite{zheng2023pointodyssey}:  median trajectory error (MTE), position accuracy ($\delta$), and survival rate.
For 3D long-term point tracking there is no prior relevant work, so we decide adapt the 2D metrics from \cite{zheng2023pointodyssey} to the 3D domain, except in terms of centimeters in 3-dimensions instead of normalized-pixels in 2D. E.g. MTE is reported as error in cm. $\delta$ is calculated at 1, 2, 4, 8 and 16cm thresholds and survival is measuring trajectories that are within 50cm of the ground-truth.

\PAR{Comparisons.}
For all three tasks of View-Synthesis, 3D tracking and 2D tracking we compare our Dynamic 3D Gaussian method to the original 3D Gaussian Splatting \cite{kerbl20233d} which we build upon. We run this in a similar online mode that we run our method for the same number of iteration steps, and thus we call this 3GS-O. 
We also perform comparisons on the dataset of Particle-NeRF \cite{abou2022particlenerf}, which includes 20 train and 10 test cameras, and contains much simpler synthetic scenes with both simple geometry and motion. On this benchmark we compare to Particle-NeRF \cite{abou2022particlenerf}, Instant-NGP \cite{mueller2022instant}, and TiNeuVox \cite{TiNeuVox}. The benchmark task for this dataset \cite{abou2022particlenerf} is defined as one in which at each timestep methods are allowed a certain number of ray renders and back-propagation steps. Since our approach doesn't work in the same way we are not able to compare exactly using these criterion. Instead as a reasonably fair criterion we run train our method for each timestep for no more wall clock time than Particle-NeRF does, e.g. 200ms per timestep.

For 3D long-term point tracking we have not found any further methods to compare against for this relatively new task. The two methods we would like most to compare to are OmniMotion \cite{wang2023tracking} adapted to multi-camera metric space and MAP Visibility Estimation \cite{joo2014map} since they seem like the most promising competitors. However, since no code is currently available for either we are unable to run such comparisons. 
For 2D tracking we compare against 2D long-term tracking approaches of which the canonical example is PIPs \cite{harley2022particle}. We provide comparison results with PIPs to show a comparison against a learnt method that was trained specifically for long-term 2D point tracking, and is also one of the state-of-the-art approaches for this task. 
Although the comparison isn't entirely fair to either method. E.g. our method sees all 27 training cameras while PIPs only sees the one that needs to be tracked in. However on the other hand PIPs was trained on 13085 training videos where ground-truth tracks were provided while our method has never seen any ground-truth tracks, nor any other video data other than the camera views for the current test scene. Regardless it is still a good test of our method to compare against such 2D point tracking methods.

\begin{figure}
    \includegraphics[width=0.7\columnwidth]{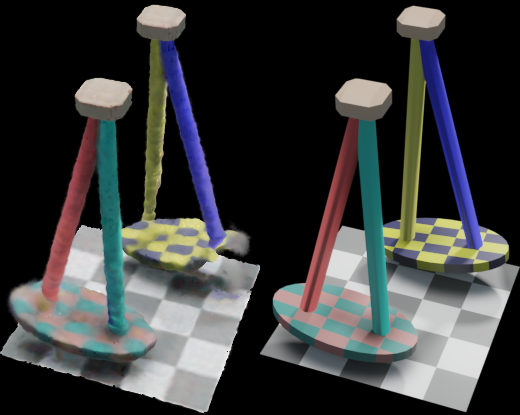}
    \caption{\textbf{Visual comparison.} Comparing Particle-NeRF (left) and  Ours (right) on the Particle-NeRF dataset.}
    \label{fig:particle}
\end{figure}

\begin{figure}
    \hspace*{-3.1cm}
    \includegraphics[width=1.36\columnwidth]{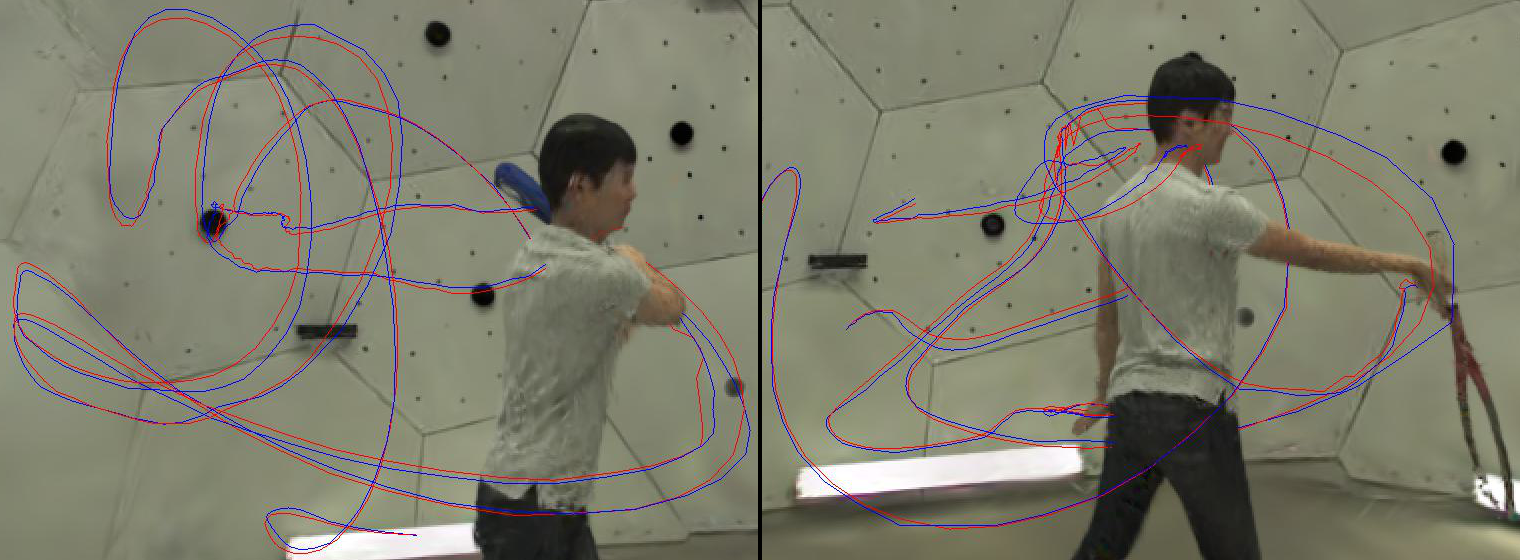}
    \caption{\textbf{Ground-truth Comparison.} Comparing our result (\textcolor{blue}{blue}) to the ground-truth (\textcolor{red}{red}). Where ground-truth is noisy, our result may be more accurate. \href{https://dynamic3dgaussians.github.io/}{[Videos]}}
    \label{fig:3dtrack}
\end{figure}

\PAR{PanopticSports Results.}
We present the results on our prepared \texttt{PanopticSports} dataset in Table~\ref{tab:cmu_panoptic}. Our approach achieves good scores on all three novel-view synthesis metrics, with a final PSNR score of 28.7. Compared to the original 3D Gaussian Splatting \cite{kerbl20233d} we achieve better PSNR and SSIM scores (although slightly worse LPIPS) across all scenes by correctly modelling the temporal consistency of the dynamic scene. Visual examples of the high-quality novel-view synthesis results can be found in Fig.~\ref{fig:teaser}.
In terms of 3D tracking, our method achieves outstanding results with a median trajectory error of only 2.21cm across all trajectories in all scenes. This is less than the width of a wrist across 150 timesteps of 3D tracking through extremely complex and fast motions (see Fig.~\ref{fig:teaser}). Our method also has 100\% survival rate across all sequences never losing the point to track and an accuracy value of 71.4. The original Gaussian Splatting \cite{kerbl20233d} approach doesn't correctly track points in 3D at all with with a much higher 55.9cm median trajectory error.
A visual comparison of our method compared to the ground-truth can be found in Fig~\ref{fig:3dtrack}.

When measuring the 2D tracking ability of our method and comparing it to the 2D tracker PIPs \cite{harley2022particle} we can see where our method really shines. Although it's not a 1-to-1 fair comparison, by comparing our numbers against the numbers of a state-of-the-art tracker we can accurately gauge the performance of our approach. We achieve a 10x lower median trajectory error of only 1.57 pixels compares to PIPs 15.7, have a much higher trajectory accuracy of 78.4 compared to 39.6, and a 100\% survival rate compared to 79\%. 

Overall these results show that our method performs excellently on the tasks of novel-view-synthesis, as well as both 2D and 3D tracking.

\begin{figure*}
    \centering
    \includegraphics[width=\linewidth]{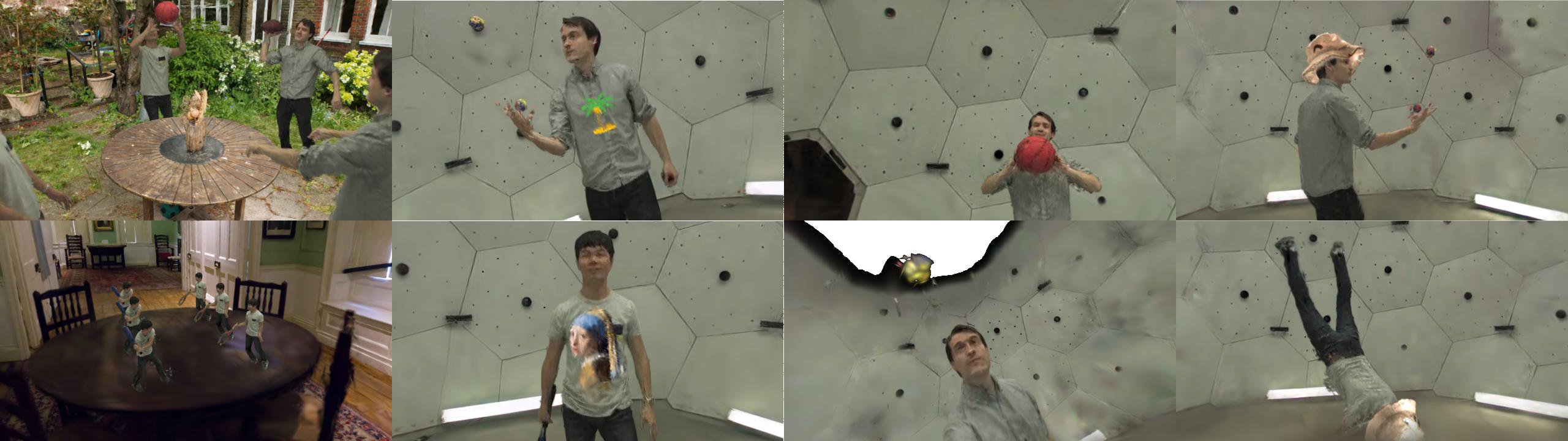}
    \caption{\textbf{Augmented reality applications enabled by Dynamic 3D Gaussians.} Left: Dynamic objects can easily be removed from scenes, duplicated and added together with other dynamic objects to new scenes. Center Left: Image edits can be lifted to 3D and then automatically propagated across time. Center Right: Camera views can be attached to dynamic Gaussians that move as the scene moves, e.g. first-person view (above) or juggling-ball's view (below). Right: Objects can be scanned and added to dynamic scenes in a way that follow the scene. E.g. the hat stays correctly on the person with the correct translation and rotation as he does a handstand. \href{https://dynamic3dgaussians.github.io/}{[Videos]}}
    \label{fig:creative}
\end{figure*}

\PAR{Particle-NeRF Dataset Results.}
On the Particle-NeRF dataset our method achieves almost perfect scores for all of PSNR, SSIM and LPIPS. This is due to the dataset being relatively simple in terms of simple synthetic objects and simple motions. A visual comparison between our method's results and those of Particle-NeRF can be seen in Fig~\ref{fig:particle}.

\PAR{Ablation Study.}
In Table~\ref{tab:ablation} we show results on the Juggle sequence of our \texttt{PanopticSports} dataset ablating the various different components of our approach. We identify 6 key components of our approach which are above and beyond that of the original 3D Gaussian splatting approach \cite{kerbl20233d}, as described in Section~\ref{sec:method}. We evaluate the effect of removing each of these components from our final method one-at-a-time as well as the effect of removing them all at once, which is then just the method from \cite{kerbl20233d} run in an online mode over different timesteps. 

For view synthesis, the original 3GS-O already works extremely well, but by correctly modelling the motion of components in the scene our full method is able to achieve a boost of 1.3 PSNR. For tracking the original doesn't accurately track the scene at all, but ours performs very accurately. In terms of key-components required for these results, all of the rigidity loss, the background segmentation loss, the colour/opacity/size parameter fixing, and the forward propagation for timestep initialization are key to obtaining both good tracking results and improvement in view-synthesis results. The rotation loss and isometric loss only provide very small improvement in the measured metrics, however visually we found the reconstruction results to be much more coherent and correct with both of these on over turning either off.

\PAR{Further Applications.}
Our Dynamic 3D Gaussian approach also leads itself nicely to being used for editing of dynamic 3D scenes. Because Gaussians are independent, subsets of them can easily be added or removed from scenes to create all sort of interesting effects. \Eg, creating realistic renders by combining multiple different dynamic components from different scenes and different backgrounds. We can also very easily propagate edits over time. \Eg, logos can be added to surfaces in a single frame, and the colors of the Gaussians can be updated to reflect this change. Such edits will automatically propagate to all other frames of a video. Finally, because we are performing full 6-DOF tracking we can take advantage of this for all sorts of visual effects, for example we could put a camera at `first person view' by attaching it to any particular Gaussian and it will follow where that Gaussian moves and rotates over time. The same can be done for adding objects to the scene that can move and rotate along with the Gaussians they are attached to. We show examples of all of these creative applications in Fig~\ref{fig:creative}.

%% file: tables/cmu_panoptic.tex
\begin{table*}[t]
\centering{}
\setlength{\tabcolsep}{1.5pt}
\renewcommand{\arraystretch}{0.95}
\footnotesize

\begin{tabular}{cccccccccccc}
\toprule 
Task & Metrics & Method &  & Juggle & Boxes & Softball & Tennis & Football & Basketball &  & Mean\tabularnewline
\midrule
\multirow{6}{*}{View Synthesis} & \multirow{2}{*}{PSNR$\uparrow$} & 3GS-O \cite{kerbl20233d} &  & 28.19 & 28.74 & \textbf{28.77} & 28.03 & \textbf{28.49} & 27.02 &  & 28.21\tabularnewline
 &  & Ours &  & \textbf{29.48} & \textbf{29.46} & 28.43 & \textbf{28.11} & \textbf{28.49} & \textbf{28.22} &  & \textbf{28.7}\tabularnewline
\cmidrule{2-12}
 & \multirow{2}{*}{SSIM$\uparrow$} & 3GS-O \cite{kerbl20233d} &  & 0.91 & \textbf{0.91} & \textbf{0.91} & 0.90 & 0.90 & 0.89 &  & 0.90\tabularnewline
 &  & Ours &  & \textbf{0.92} & \textbf{0.91} & \textbf{0.91} & \textbf{0.91} & \textbf{0.91} & \textbf{0.91} &  & \textbf{0.91}\tabularnewline
\cmidrule{2-12}
 & \multirow{2}{*}{LPIPS$\downarrow$} & 3GS-O \cite{kerbl20233d} &  & \textbf{0.15} & \textbf{0.15} & \textbf{0.14} & \textbf{0.16} & \textbf{0.16} & \textbf{0.18} &  & \textbf{0.16}\tabularnewline
 &  & Ours &  & \textbf{0.15} & 0.17 & 0.19 & 0.17 & 0.19 & \textbf{0.18} &  & \textbf{0.17}\tabularnewline
\midrule
\multirow{6}{*}{3D Tracking} & \multirow{2}{*}{3D MTE$\downarrow$} & 3GS-O \cite{kerbl20233d} &  & 32.81 & 39.95 & 64.94 & 75.54 & 45.57 & 76.71 &  & 55.9\tabularnewline
 &  & Ours &  & \textbf{1.90} & \textbf{1.97} & \textbf{2.02} & \textbf{2.33} & \textbf{2.45} & \textbf{2.56} &  & \textbf{2.21}\tabularnewline
\cmidrule{2-12}
 & \multirow{2}{*}{3D $\delta$$\uparrow$} & 3GS-O \cite{kerbl20233d} &  & 13.6 & 3.5 & 5.9 & 4.2 & 9.8 & 3.5 &  & 6.8\tabularnewline
 &  & Ours &  & \textbf{77.2} & \textbf{75.9} & \textbf{70.3} & \textbf{69.0} & \textbf{69.4} & \textbf{66.3} &  & \textbf{71.4}\tabularnewline
\cmidrule{2-12} \cmidrule{11-12} \cmidrule{12-12} 
 & \multirow{2}{*}{3D Surv$\uparrow$} & 3GS-O \cite{kerbl20233d} &  & 56.3 & 60.8 & 37.2 & 16.9 & 59.6 & 31.9 &  & 43.8\tabularnewline
 &  & Ours &  & \textbf{100} & \textbf{100} & \textbf{100} & \textbf{100} & \textbf{100} & \textbf{100} &  & \textbf{100}\tabularnewline
\midrule
\multirow{9}{*}{2D Tracking} & \multirow{3}{*}{2D MTE$\downarrow$} & 3GS-O \cite{kerbl20233d} &  & 23.86 & 29.88 & 51.6 & 58.15 & 35.15 & 64.29 &  & 43.8\tabularnewline
 &  & PIPS \cite{harley2022particle} &  & 5.76 & 8.42 & 13.3 & 21.0 & 23.2 & 22.6 &  & 15.7\tabularnewline
 &  & Ours &  & \textbf{1.54} & \textbf{1.42} & \textbf{1.69} & \textbf{1.36} & \textbf{1.48} & \textbf{1.93} &  & \textbf{1.57}\tabularnewline
\cmidrule{2-12}
 & \multirow{3}{*}{2D $\delta$$\uparrow$} & 3GS-O \cite{kerbl20233d} &  & 17.1 & 10.5 & 8.9 & 6.5 & 15.0 & 7.2 &  & 10.9\tabularnewline
 &  & PIPS \cite{harley2022particle} &  & 55.9 & 39.5 & 37.0 & 28.4 & 43.5 & 33.2 &  & 39.6\tabularnewline
 &  & Ours &  & \textbf{80.4} & \textbf{82.5} & \textbf{77.3} & \textbf{80.2} & \textbf{79.7} & \textbf{73.9} &  & \textbf{78.4}\tabularnewline
\cmidrule{2-12}
 & \multirow{3}{*}{2D Surv$\uparrow$} & 3GS-O \cite{kerbl20233d} &  & 71.3 & 74.4 & 42.7 & 23.0 & 69.6 & 47.1 &  & 54.7\tabularnewline
 &  & PIPS \cite{harley2022particle} &  & 91.6 & 61.3 & 88.6 & 72.2 & 79.8 & 77.6 &  & 79.0\tabularnewline
 &  & Ours &  & \textbf{100} & \textbf{100} & \textbf{100} & \textbf{100} & \textbf{100} & \textbf{100} &  & \textbf{100}\tabularnewline
\bottomrule
\end{tabular}

\caption{\label{tab:cmu_panoptic}{\textbf{Results on our prepared PanopticSports dataset}. See text for details on the dataset, metrics, tasks and methods.}}
\end{table*}

%% file: tables/particle-nerf.tex
\begin{table}[t]
\centering{}
\setlength{\tabcolsep}{2.5pt}
\footnotesize

\begin{tabular}{ccccc}
\toprule 
Method &  & PSNR$\uparrow$ & SSIM$\uparrow$ & LPIPS$\downarrow$\tabularnewline
\midrule 
TiNeuVox-S \cite{TiNeuVox} &  & 26.64 & 0.92 & 0.14\tabularnewline
TiNeuVox \cite{TiNeuVox} &  & 27.28 & 0.91 & 0.13\tabularnewline
InstantNGP \cite{mueller2022instant} &  & 24.69 & 0.91 & 0.12\tabularnewline
Particle-NeRF \cite{abou2022particlenerf} &  & 27.47 & 0.94 & 0.08 \tabularnewline
Ours &  & \textbf{39.49} & \textbf{0.99} & \textbf{0.02}\tabularnewline
\bottomrule
\end{tabular}

\caption{\label{tab:particle}{\textbf{Result on the Particle-NeRF dataset}. See text for details on the dataset, metrics, tasks and methods.}}
\end{table}

%% file: tables/ablations.tex
\begin{table*}[t]
\centering{}
\setlength{\tabcolsep}{2.0pt}
\footnotesize

\begin{tabular}{cccccccccccccccccc}
\toprule 
\multirow{2}{*}{Exp \#} & \multirow{2}{*}{Description} & \multicolumn{6}{c}{Additions} &  & \multicolumn{3}{c}{View Synthesis} &  & \multicolumn{2}{c}{3D Tracking} &  & \multicolumn{2}{c}{2D Tracking}\tabularnewline
 &  & $\mathcal{L}_{\text{\text{{Rigid}}}}$ & $\mathcal{L}_{\text{\text{{Rot}}}}$ & $\mathcal{L}_{\text{\text{{Iso}}}}$ & $\mathcal{L}_{\text{\text{{Bg}}}}$ & Fix & Prop &  & PSNR$\uparrow$ & SSIM$\uparrow$ & LPIPS$\downarrow$ &  & 3D MTE$\downarrow$ & 3D $\delta$$\uparrow$ &  & 2D MTE$\downarrow$ & 2D $\delta$$\uparrow$\tabularnewline
\midrule
0 & Ours - Full & \cmark & \cmark & \cmark & \cmark & \cmark & \cmark &  & \textbf{29.48} & \textbf{0.92} & \textbf{0.15} &  & \textbf{1.90} & \textbf{77.2} &  & \textbf{1.54} & \textbf{80.4}\tabularnewline
1 & No $\mathcal{L}_{\text{\text{{Rigid}}}}$ & \xmark & \cmark & \cmark & \cmark & \cmark & \cmark &  & 28.51 & 0.91 & 0.17 &  & 4.32 & 55.2 &  & 3.80 & 58.7\tabularnewline
2 & No $\mathcal{L}_{\text{\text{{Rot}}}}$ & \cmark & \xmark & \cmark & \cmark & \cmark & \cmark &  & 29.43 & \textbf{0.92} & 0.16 &  & 1.91 & 76.6 &  & 1.55 & 79.8\tabularnewline
3 & No $\mathcal{L}_{\text{\text{{Iso}}}}$ & \cmark & \cmark & \xmark & \cmark & \cmark & \cmark &  & 29.36 & \textbf{0.92} & 0.16 &  & 1.93 & 76.7 &  & 1.72 & 79.3\tabularnewline
4 & No $\mathcal{L}_{\text{\text{{Bg}}}}$ & \cmark & \cmark & \cmark & \xmark & \cmark & \cmark &  & 24.14 & 0.82 & 0.34 &  & 8.46 & 60.0 &  & 6.40 & 63.2\tabularnewline
5 & No Param Fixing & \cmark & \cmark & \cmark & \cmark & \xmark & \cmark &  & 27.14 & 0.89 & 0.22 &  & 30.7 & 57.7 &  & 19.15 & 58.8\tabularnewline
6 & No Forward Prop & \cmark & \cmark & \cmark & \cmark & \cmark & \xmark &  & 28.48 & 0.91 & 0.16 &  & 6.32 & 54.87 &  & 5.4 & 57.7\tabularnewline
7 & 3GS-O \cite{kerbl20233d} & \xmark & \xmark & \xmark & \xmark & \xmark & \xmark &  & 28.19 & 0.90 & \textbf{0.15} &  & 32.81 & 13.6 &  & 23.86 & 17.1\tabularnewline
\bottomrule
\end{tabular}

\caption{\label{tab:ablation}{\textbf{Ablation results on the Juggle scene of PanopticSports.} See text for details on the dataset, metrics, tasks and methods.}}
\end{table*}

%% file: sec/5_conclusion.tex
\vspace{-6pt}
\section{Conclusion and Limitations}
\vspace{-3pt}
\label{sec:conclusion}

\PAR{Limitations}
While our method achieves excellent results it is not without limitations. For example, by design our method is only able to track parts of scenes that are visible in the initial frame. It would completely fail to reconstruct new objects entering the scene. Our method also requires a multi-camera setup and does not work off-the-shelf on monocular video. We believe that these limitations are the seeds of exciting future research directions to build upon and extend our Dynamic 3D Gaussian representation. 

\PAR{Conclusion}
In this work, we have introduced a novel method for dynamic 3D scene modeling, view synthesis, and 6-DOF tracking that has relevant applications across various domains, including entertainment, robotics, VR and AR. Utilizing Gaussian elements to model dynamic scenes, our approach uniquely captures movements and rotations, consistent with physical properties. The implications of our method extend beyond the immediate results, offering new avenues for real-time rendering and creative scene editing. Our approach, characterized by efficiency and accuracy, sets a promising direction for future research and practical applications in 3D modeling and tracking, underscoring the potential for further innovation in these fields.

{\footnotesize \PAR{Acknowledgements:} Jonathon Luiten's research was funded, in parts, by NRW Verbundprojekt WestAI (01IS22094D).
EP/N019474/1. The authors would like to thank Joanna Materzynska, Leonid Keselman, Dinesh Reddy, Jonas Schult and Adam Harley for helpful discussions.}

\newpage

%% file: main.bbl
\begin{thebibliography}{48}
\providecommand{\natexlab}[1]{#1}
\providecommand{\url}[1]{\texttt{#1}}
\expandafter\ifx\csname urlstyle\endcsname\relax
  \providecommand{\doi}[1]{doi: #1}\else
  \providecommand{\doi}{doi: \begingroup \urlstyle{rm}\Url}\fi

\bibitem[Abou-Chakra et~al.(2022)Abou-Chakra, Dayoub, and
  S{\"u}nderhauf]{abou2022particlenerf}
Jad Abou-Chakra, Feras Dayoub, and Niko S{\"u}nderhauf.
\newblock Particlenerf: Particle based encoding for online neural radiance
  fields in dynamic scenes.
\newblock \emph{arXiv:2211.04041}, 2022.

\bibitem[Bansal and Zollhoefer(2023)]{bansal2023neural}
Aayush Bansal and Michael Zollhoefer.
\newblock Neural pixel composition for 3d-4d view synthesis from multi-views.
\newblock In \emph{CVPR}, pages 290--299, 2023.

\bibitem[B{\^a}rsan et~al.(2018)B{\^a}rsan, Liu, Pollefeys, and
  Geiger]{barsan2018robust}
Ioan~Andrei B{\^a}rsan, Peidong Liu, Marc Pollefeys, and Andreas Geiger.
\newblock Robust dense mapping for large-scale dynamic environments.
\newblock In \emph{ICRA}, pages 7510--7517. IEEE, 2018.

\bibitem[Beauchemin and Barron(1995)]{beauchemin1995computation}
Steven~S. Beauchemin and John~L. Barron.
\newblock The computation of optical flow.
\newblock \emph{ACM computing surveys (CSUR)}, 27\penalty0 (3):\penalty0
  433--466, 1995.

\bibitem[Cao and Johnson(2023)]{cao2023hexplane}
Ang Cao and Justin Johnson.
\newblock Hexplane: A fast representation for dynamic scenes.
\newblock In \emph{CVPR}, pages 130--141, 2023.

\bibitem[Doersch et~al.(2022)Doersch, Gupta, Markeeva, Continente, Smaira,
  Aytar, Carreira, Zisserman, and Yang]{doersch2022tap}
Carl Doersch, Ankush Gupta, Larisa Markeeva, Adria~Recasens Continente, Kucas
  Smaira, Yusuf Aytar, Joao Carreira, Andrew Zisserman, and Yi Yang.
\newblock Tap-vid: A benchmark for tracking any point in a video.
\newblock In \emph{NeurIPS Datasets Track}, 2022.

\bibitem[Doersch et~al.(2023)Doersch, Yang, Vecerik, Gokay, Gupta, Aytar,
  Carreira, and Zisserman]{doersch2023tapir}
Carl Doersch, Yi Yang, Mel Vecerik, Dilara Gokay, Ankush Gupta, Yusuf Aytar,
  Joao Carreira, and Andrew Zisserman.
\newblock Tapir: Tracking any point with per-frame initialization and temporal
  refinement.
\newblock \emph{arXiv:2306.08637}, 2023.

\bibitem[Du et~al.(2021)Du, Zhang, Yu, Tenenbaum, and Wu]{du2021neural}
Yilun Du, Yinan Zhang, Hong-Xing Yu, Joshua~B Tenenbaum, and Jiajun Wu.
\newblock Neural radiance flow for 4d view synthesis and video processing.
\newblock In \emph{ICCV}, pages 14304--14314. IEEE Computer Society, 2021.

\bibitem[Fang et~al.(2022)Fang, Yi, Wang, Xie, Zhang, Liu, Nie\ss{}ner, and
  Tian]{TiNeuVox}
Jiemin Fang, Taoran Yi, Xinggang Wang, Lingxi Xie, Xiaopeng Zhang, Wenyu Liu,
  Matthias Nie\ss{}ner, and Qi Tian.
\newblock Fast dynamic radiance fields with time-aware neural voxels.
\newblock In \emph{SIGGRAPH Asia 2022 Conference Papers}, 2022.

\bibitem[Fridovich-Keil et~al.(2023)Fridovich-Keil, Meanti, Warburg, Recht, and
  Kanazawa]{fridovich2023k}
Sara Fridovich-Keil, Giacomo Meanti, Frederik~Rahb{\ae}k Warburg, Benjamin
  Recht, and Angjoo Kanazawa.
\newblock K-planes: Explicit radiance fields in space, time, and appearance.
\newblock In \emph{CVPR}, pages 12479--12488, 2023.

\bibitem[Gao et~al.(2022)Gao, Li, Tulsiani, Russell, and
  Kanazawa]{gao2022monocular}
Hang Gao, Ruilong Li, Shubham Tulsiani, Bryan Russell, and Angjoo Kanazawa.
\newblock Monocular dynamic view synthesis: A reality check.
\newblock \emph{Advances in Neural Information Processing Systems},
  35:\penalty0 33768--33780, 2022.

\bibitem[Harley et~al.(2022)Harley, Fang, and Fragkiadaki]{harley2022particle}
Adam~W. Harley, Zhaoyuan Fang, and Katerina Fragkiadaki.
\newblock Particle video revisited: {T}racking through occlusions using point
  trajectories.
\newblock In \emph{ECCV}, 2022.

\bibitem[I\c{s}{\i}k et~al.(2023)I\c{s}{\i}k, Rünz, Georgopoulos, Khakhulin,
  Starck, Agapito, and Nießner]{isik2023humanrf}
Mustafa I\c{s}{\i}k, Martin Rünz, Markos Georgopoulos, Taras Khakhulin,
  Jonathan Starck, Lourdes Agapito, and Matthias Nießner.
\newblock Humanrf: High-fidelity neural radiance fields for humans in motion.
\newblock \emph{ACM Transactions on Graphics (TOG)}, 42\penalty0 (4):\penalty0
  1--12, 2023.

\bibitem[Joo et~al.(2014)Joo, Soo~Park, and Sheikh]{joo2014map}
Hanbyul Joo, Hyun Soo~Park, and Yaser Sheikh.
\newblock Map visibility estimation for large-scale dynamic 3d reconstruction.
\newblock In \emph{CVPR}, pages 1122--1129, 2014.

\bibitem[Joo et~al.(2015)Joo, Liu, Tan, Gui, Nabbe, Matthews, Kanade, Nobuhara,
  and Sheikh]{joo2015panoptic}
Hanbyul Joo, Hao Liu, Lei Tan, Lin Gui, Bart Nabbe, Iain Matthews, Takeo
  Kanade, Shohei Nobuhara, and Yaser Sheikh.
\newblock Panoptic studio: A massively multiview system for social motion
  capture.
\newblock In \emph{Proceedings of the IEEE International Conference on Computer
  Vision}, pages 3334--3342, 2015.

\bibitem[Karaev et~al.(2023)Karaev, Rocco, Graham, Neverova, Vedaldi, and
  Rupprecht]{karaev2023cotracker}
Nikita Karaev, Ignacio Rocco, Benjamin Graham, Natalia Neverova, Andrea
  Vedaldi, and Christian Rupprecht.
\newblock Cotracker: It is better to track together.
\newblock \emph{arXiv:2307.07635}, 2023.

\bibitem[Kerbl et~al.(2023)Kerbl, Kopanas, Leimkuehler, and
  Drettakis]{kerbl20233d}
Bernhard Kerbl, Georgios Kopanas, Thomas Leimkuehler, and George Drettakis.
\newblock 3d gaussian splatting for real-time radiance field rendering.
\newblock \emph{ACM Transactions on Graphics (TOG)}, 42\penalty0 (4):\penalty0
  1--14, 2023.

\bibitem[Keselman and Hebert(2022)]{keselman2022fuzzy}
Leonid Keselman and Martial Hebert.
\newblock Approximate differentiable rendering with algebraic surfaces.
\newblock In \emph{ECCV}, 2022.

\bibitem[Kristan et~al.(2021)Kristan, Matas, Leonardis, Felsberg, Pflugfelder,
  K\"am\"ar\"ainen, Chang, Danelljan, Cehovin, Luke\v{z}i\v{c}, Drbohlav,
  K\"apyl\"a, H\"ager, Yan, Yang, Zhang, and Fern\'andez]{Kristan_2021_ICCV}
Matej Kristan, Ji\v{r}{\'\i} Matas, Ale\v{s} Leonardis, Michael Felsberg, Roman
  Pflugfelder, Joni-Kristian K\"am\"ar\"ainen, Hyung~Jin Chang, Martin
  Danelljan, Luka Cehovin, Alan Luke\v{z}i\v{c}, Ondrej Drbohlav, Jani
  K\"apyl\"a, Gustav H\"ager, Song Yan, Jinyu Yang, Zhongqun Zhang, and Gustavo
  Fern\'andez.
\newblock The ninth visual object tracking vot2021 challenge results.
\newblock In \emph{ICCVW}, pages 2711--2738, 2021.

\bibitem[Li et~al.(2022)Li, Tanke, Vo, Zollhofer, Gall, Kanazawa, and
  Lassner]{li2022tava}
Ruilong Li, Julian Tanke, Minh Vo, Michael Zollhofer, Jurgen Gall, Angjoo
  Kanazawa, and Christoph Lassner.
\newblock Tava: Template-free animatable volumetric actors.
\newblock In \emph{ECCV}, 2022.

\bibitem[Li et~al.(2021)Li, Niklaus, Snavely, and Wang]{li2021neural}
Zhengqi Li, Simon Niklaus, Noah Snavely, and Oliver Wang.
\newblock Neural scene flow fields for space-time view synthesis of dynamic
  scenes.
\newblock In \emph{CVPR}, pages 6498--6508, 2021.

\bibitem[Liu et~al.(2023)Liu, Gao, Meuleman, Tseng, Saraf, Kim, Chuang, Kopf,
  and Huang]{liu2023robust}
Yu-Lun Liu, Chen Gao, Andreas Meuleman, Hung-Yu Tseng, Ayush Saraf, Changil
  Kim, Yung-Yu Chuang, Johannes Kopf, and Jia-Bin Huang.
\newblock Robust dynamic radiance fields.
\newblock In \emph{CVPR}, pages 13--23, 2023.

\bibitem[Luiten et~al.(2020)Luiten, Fischer, and Leibe]{luiten2019track}
Jonathon Luiten, Tobias Fischer, and Bastian Leibe.
\newblock Track to reconstruct and reconstruct to track.
\newblock \emph{IEEE Robotics and Automation Letters}, 5\penalty0 (2):\penalty0
  1803--1810, 2020.

\bibitem[Max(1995)]{max1995optical}
Nelson Max.
\newblock Optical models for direct volume rendering.
\newblock \emph{IEEE Transactions on Visualization and Computer Graphics},
  1\penalty0 (2):\penalty0 99--108, 1995.

\bibitem[Mildenhall et~al.(2020)Mildenhall, Srinivasan, Tancik, Barron,
  Ramamoorthi, and Ng]{mildenhall2020nerf}
Ben Mildenhall, Pratul~P. Srinivasan, Matthew Tancik, Jonathan~T. Barron, Ravi
  Ramamoorthi, and Ren Ng.
\newblock Nerf: Representing scenes as neural radiance fields for view
  synthesis.
\newblock In \emph{ECCV}, 2020.

\bibitem[M\"uller et~al.(2022)M\"uller, Evans, Schied, and
  Keller]{mueller2022instant}
Thomas M\"uller, Alex Evans, Christoph Schied, and Alexander Keller.
\newblock Instant neural graphics primitives with a multiresolution hash
  encoding.
\newblock \emph{ACM Trans. Graph.}, 41\penalty0 (4):\penalty0 102:1--102:15,
  2022.

\bibitem[Newcombe et~al.(2015)Newcombe, Fox, and
  Seitz]{newcombe2015dynamicfusion}
Richard~A Newcombe, Dieter Fox, and Steven~M Seitz.
\newblock Dynamicfusion: Reconstruction and tracking of non-rigid scenes in
  real-time.
\newblock In \emph{CVPR}, pages 343--352, 2015.

\bibitem[Park et~al.(2021{\natexlab{a}})Park, Sinha, Barron, Bouaziz, Goldman,
  Seitz, and Martin-Brualla]{park2021nerfies}
Keunhong Park, Utkarsh Sinha, Jonathan~T Barron, Sofien Bouaziz, Dan~B Goldman,
  Steven~M Seitz, and Ricardo Martin-Brualla.
\newblock Nerfies: Deformable neural radiance fields.
\newblock In \emph{ICCV}, pages 5865--5874, 2021{\natexlab{a}}.

\bibitem[Park et~al.(2021{\natexlab{b}})Park, Sinha, Hedman, Barron, Bouaziz,
  Goldman, Martin-Brualla, and Seitz]{park2021hypernerf}
Keunhong Park, Utkarsh Sinha, Peter Hedman, Jonathan~T. Barron, Sofien Bouaziz,
  Dan~B Goldman, Ricardo Martin-Brualla, and Steven~M. Seitz.
\newblock Hypernerf: A higher-dimensional representation for topologically
  varying neural radiance fields.
\newblock \emph{ACM Trans. Graph.}, 40\penalty0 (6), 2021{\natexlab{b}}.

\bibitem[Perazzi et~al.(2016)Perazzi, Pont-Tuset, McWilliams, {Van Gool},
  Gross, and Sorkine-Hornung]{Caelles_arXiv_2019}
F. Perazzi, J. Pont-Tuset, B. McWilliams, L. {Van Gool}, M. Gross, and A.
  Sorkine-Hornung.
\newblock A benchmark dataset and evaluation methodology for video object
  segmentation.
\newblock In \emph{CVPR}, 2016.

\bibitem[Prokudin et~al.(2023)Prokudin, Ma, Raafat, Valentin, and
  Tang]{prokudin2023dynamic}
Sergey Prokudin, Qianli Ma, Maxime Raafat, Julien Valentin, and Siyu Tang.
\newblock Dynamic point fields.
\newblock \emph{arXiv:2304.02626}, 2023.

\bibitem[Pumarola et~al.(2021)Pumarola, Corona, Pons-Moll, and
  Moreno-Noguer]{pumarola2021d}
Albert Pumarola, Enric Corona, Gerard Pons-Moll, and Francesc Moreno-Noguer.
\newblock D-nerf: Neural radiance fields for dynamic scenes.
\newblock In \emph{CVPR}, pages 10318--10327, 2021.

\bibitem[Simon et~al.(2017)Simon, Joo, and Sheikh]{Simon_2017_CVPR}
Tomas Simon, Hanbyul Joo, and Yaser Sheikh.
\newblock Hand keypoint detection in single images using multiview
  bootstrapping.
\newblock \emph{CVPR}, 2017.

\bibitem[Slavcheva et~al.(2017)Slavcheva, Baust, Cremers, and
  Ilic]{slavcheva2017killingfusion}
Miroslava Slavcheva, Maximilian Baust, Daniel Cremers, and Slobodan Ilic.
\newblock Killingfusion: Non-rigid 3d reconstruction without correspondences.
\newblock In \emph{CVPR}, pages 1386--1395, 2017.

\bibitem[Song et~al.(2023)Song, Yang, Deng, Zhu, and Ramanan]{song2023total}
Chonghyuk Song, Gengshan Yang, Kangle Deng, Jun-Yan Zhu, and Deva Ramanan.
\newblock Total-recon: Deformable scene reconstruction for embodied view
  synthesis.
\newblock In \emph{ICCV}, 2023.

\bibitem[Turki et~al.(2023)Turki, Zhang, Ferroni, and Ramanan]{turki2023suds}
Haithem Turki, Jason~Y Zhang, Francesco Ferroni, and Deva Ramanan.
\newblock Suds: Scalable urban dynamic scenes.
\newblock In \emph{CVPR}, pages 12375--12385, 2023.

\bibitem[Vedula et~al.(1999)Vedula, Baker, Rander, Collins, and
  Kanade]{vedula1999three}
Sundar Vedula, Simon Baker, Peter Rander, Robert Collins, and Takeo Kanade.
\newblock Three-dimensional scene flow.
\newblock In \emph{Proceedings of the Seventh IEEE International Conference on
  Computer Vision}, pages 722--729. IEEE, 1999.

\bibitem[Voigtlaender et~al.(2019)Voigtlaender, Krause, Osep, Luiten, Sekar,
  Geiger, and Leibe]{voigtlaender2019mots}
Paul Voigtlaender, Michael Krause, Aljosa Osep, Jonathon Luiten, Berin
  Balachandar~Gnana Sekar, Andreas Geiger, and Bastian Leibe.
\newblock Mots: Multi-object tracking and segmentation.
\newblock In \emph{CVPR}, pages 7942--7951, 2019.

\bibitem[Wang et~al.(2022)Wang, Wang, Sun, Kortylewski, and
  Yuille]{wang2022voge}
Angtian Wang, Peng Wang, Jian Sun, Adam Kortylewski, and Alan Yuille.
\newblock Voge: a differentiable volume renderer using gaussian ellipsoids for
  analysis-by-synthesis.
\newblock In \emph{ICLR}, 2022.

\bibitem[Wang et~al.(2023)Wang, Chang, Cai, Li, Hariharan, Holynski, and
  Snavely]{wang2023tracking}
Qianqian Wang, Yen-Yu Chang, Ruojin Cai, Zhengqi Li, Bharath Hariharan,
  Aleksander Holynski, and Noah Snavely.
\newblock Tracking everything everywhere all at once.
\newblock \emph{arXiv:2306.05422}, 2023.

\bibitem[Wang et~al.(2004)Wang, Bovik, Sheikh, and Simoncelli]{wang2004image}
Zhou Wang, Alan~C Bovik, Hamid~R Sheikh, and Eero~P Simoncelli.
\newblock Image quality assessment: from error visibility to structural
  similarity.
\newblock \emph{IEEE transactions on image processing}, 13\penalty0
  (4):\penalty0 600--612, 2004.

\bibitem[Weng et~al.(2022)Weng, Curless, Srinivasan, Barron, and
  Kemelmacher-Shlizerman]{weng2022humannerf}
Chung-Yi Weng, Brian Curless, Pratul~P Srinivasan, Jonathan~T Barron, and Ira
  Kemelmacher-Shlizerman.
\newblock Humannerf: Free-viewpoint rendering of moving people from monocular
  video.
\newblock In \emph{CVPR}, pages 16210--16220, 2022.

\bibitem[Xian et~al.(2021)Xian, Huang, Kopf, and Kim]{xian2020space}
Wenqi Xian, Jia-Bin Huang, Johannes Kopf, and Changil Kim.
\newblock Space-time neural irradiance fields for free-viewpoint video.
\newblock \emph{CVPR}, 2021.

\bibitem[Yang et~al.(2022)Yang, Vo, Neverova, Ramanan, Vedaldi, and
  Joo]{yang2022banmo}
Gengshan Yang, Minh Vo, Natalia Neverova, Deva Ramanan, Andrea Vedaldi, and
  Hanbyul Joo.
\newblock Banmo: Building animatable 3d neural models from many casual videos.
\newblock In \emph{CVPR}, pages 2863--2873, 2022.

\bibitem[Zhang et~al.(2022)Zhang, Baek, Rusinkiewicz, and
  Heide]{zhang2022differentiable}
Qiang Zhang, Seung-Hwan Baek, Szymon Rusinkiewicz, and Felix Heide.
\newblock Differentiable point-based radiance fields for efficient view
  synthesis.
\newblock In \emph{SIGGRAPH Asia 2022 Conference Papers}, pages 1--12, 2022.

\bibitem[Zhang et~al.(2018)Zhang, Isola, Efros, Shechtman, and
  Wang]{zhang2018perceptual}
Richard Zhang, Phillip Isola, Alexei~A Efros, Eli Shechtman, and Oliver Wang.
\newblock The unreasonable effectiveness of deep features as a perceptual
  metric.
\newblock In \emph{CVPR}, 2018.

\bibitem[Zheng et~al.(2023)Zheng, Harley, Shen, Wetzstein, and
  Guibas]{zheng2023pointodyssey}
Yang Zheng, Adam~W. Harley, Bokui Shen, Gordon Wetzstein, and Leonidas~J.
  Guibas.
\newblock Pointodyssey: A large-scale synthetic dataset for long-term point
  tracking.
\newblock In \emph{ICCV}, 2023.

\bibitem[Zwicker et~al.(2001)Zwicker, Pfister, Van~Baar, and
  Gross]{zwicker2001surface}
Matthias Zwicker, Hanspeter Pfister, Jeroen Van~Baar, and Markus Gross.
\newblock Surface splatting.
\newblock In \emph{Proceedings of the 28th annual conference on Computer
  graphics and interactive techniques}, pages 371--378, 2001.

\end{thebibliography}
